\documentclass[10pt,conference]{IEEEtran}
\IEEEoverridecommandlockouts

\usepackage{graphicx}
\usepackage{amsmath}
\usepackage{multirow}
\usepackage{cite}
\usepackage{url}
\usepackage{booktabs}

\begin{document}

\title{NeuroAPS-Net: Neuro-Anatomically Aware Point Cloud Representation for Efficient Alzheimer's Disease Classification}

\author{
\IEEEauthorblockN{Towhidul Islam\IEEEauthorrefmark{1}, Mufti Mahmud\IEEEauthorrefmark{1}\IEEEauthorrefmark{2}\IEEEauthorrefmark{3}}

\IEEEauthorblockA{\IEEEauthorrefmark{1}
ICS Department, King Fahd University of Petroleum \& Minerals, Dhahran, Saudi Arabia}

\IEEEauthorblockA{\IEEEauthorrefmark{2}
SDAIA-KFUPM JRC for AI, King Fahd University of Petroleum \& Minerals, Dhahran, Saudi Arabia}

\IEEEauthorblockA{\IEEEauthorrefmark{3}
IRC for Bio Systems and Machines, King Fahd University of Petroleum \& Minerals, Dhahran, Saudi Arabia}

\IEEEauthorblockA{
Emails: g202416880@kfupm.edu.sa, towhidulislam133086@gmail.com, mufti.mahmud@kfupm.edu.sa, muftimahmud@gmail.com}
}

\maketitle

\begin{abstract}
Alzheimer's disease (AD) is a progressive neurodegenerative disorder and a major cause of dementia. Structural MRI is widely used to analyse AD-related brain atrophy; however, most deep learning methods rely on computationally expensive 3D convolutional neural networks (CNNs), limiting deployment in resource-constrained settings. This work introduces two main contributions. First, we propose a pipeline that converts T1-weighted MRI into anatomically informed 2D point clouds using Anatomical Priority Sampling (APS), producing ADNI-2DPC, the first neuroanatomically labelled MRI-derived point cloud dataset. Second, we present NeuroAPS-Net, a lightweight geometric deep learning model that incorporates anatomical priors via region-aware feature encoding and ROI token aggregation. Experiments on ADNI-2DPC demonstrate that NeuroAPS-Net achieves competitive classification accuracy while significantly reducing inference latency and GPU memory compared to state-of-the-art point cloud methods. These results highlight the potential of anatomically guided point cloud learning as an efficient and interpretable alternative to voxel-based CNNs for AD classification.

\end{abstract}


\section{Introduction}

Alzheimer's disease (AD) is a progressive neurodegenerative disorder and a leading cause of dementia in people over 65. Approximately one in nine people in this age group is affected, making early and precise diagnosis a major clinical challenge \cite{wang2026classification}. Structural MRI is widely used in AD diagnosis by analysing atrophy in regions such as the hippocampus, ventricles, and cortex, which are linked to disease progression. \cite{Hechkel2025Unveiling}. Recent deep learning methods based on 3D convolutional neural networks (CNNs) achieve strong performance in MRI-based AD classification \cite{IJCNN1}. However, they rely on dense voxel representations, resulting in high computational cost, memory usage, and latency, which limit deployment in resource-constrained clinical settings \cite{pang2025slim}.

Point cloud representations offer a compact alternative to voxel grids by modelling structures as discrete points, enabling efficient geometric learning \cite{IJCNN1}. Despite their success in computer vision, their application to neuroimaging remains limited, particularly in incorporating disease-specific anatomical priors and providing MRI-derived point cloud datasets.

This work addresses how MRI can be efficiently represented as point clouds while preserving clinically relevant anatomical information for AD classification. By leveraging anatomically guided sampling and lightweight model design, the proposed approach aims to provide an efficient alternative to voxel-based methods, suitable for deployment in resource-constrained clinical environments. To address these challenges, this paper makes two core contributions: 

\begin{itemize}
    \item First, we propose a novel pipeline that converts T1-weighted MRI into anatomically informed 2D point clouds using Anatomical Priority Sampling (APS), emphasising disease-relevant regions and producing the ADNI-2DPC dataset.

    \item Second, we introduce NeuroAPS-Net, a lightweight geometric deep learning model with ROI-based feature aggregation for efficient and interpretable AD classification, achieving lower computational cost compared to voxel-based and existing point cloud models.
\end{itemize}

Unlike conventional point cloud methods that treat all points uniformly, the proposed framework integrates neuroanatomical priors into both representation and learning, enabling region-aware feature modelling tailored to Alzheimer’s disease.

\section{Related Work}
\label{sec-relwork}
Recent advancements in Alzheimer’s disease classification fall into three main research directions: voxel-based deep learning on MRI, point cloud–based geometric deep learning for computer vision tasks, and emerging medical applications of point cloud representations. This section reviews current work in each direction and addresses the limitations that motivate the proposed methodology.

\textit{Voxel-Based Deep Learning for Neuroimaging:} Recent voxel-based approaches (e.g., HemiNet \cite{liu2026leveraging} and CNN-based models \cite{das2026cnn}) have demonstrated strong performance in MRI-based Alzheimer’s disease classification. These approaches operate on dense volumetric data and can capture complex spatial patterns. However, they rely on computationally intensive processing, require extensive preprocessing, and offer limited explicit modelling of anatomical geometry, which limits their applicability in resource-constrained clinical settings.

\textit{Point Clouds in Computer Vision Tasks:} Recent geometric deep learning approaches (e.g., RenCAD \cite{lu2026autoregressive} and PU-Mamba \cite{li2026pu}) highlight the representational strength of point cloud learning using transformer-based and sequence modelling techniques. In addition, widely used architectures such as PointNet, PointNet++, DGCNN, PointCNN, and Point Transformer have significantly advanced 3D representation learning for generic objects and scenes.

However, these methods are primarily designed for computer vision applications and often rely on computationally intensive architectures. Moreover, these methods assume uniform importance across points and do not incorporate domain-specific anatomical knowledge. This limits their effectiveness for efficient, interpretable, and disease-specific neuroimaging tasks.

\textit{Point Clouds in Medical Imaging:} Recent studies have explored point cloud representations in medical imaging across various tasks. Atici et al. (2026) proposed a transformer-based framework for estimating internal organ geometry from body-surface point clouds, demonstrating the ability of sparse representations to capture anatomical structures without voxelization \cite{atici2026surface}. Mekhzoum et al. (2025) introduced a point cloud–based registration approach for dynamic 4D-CT imaging, achieving competitive accuracy with reduced computational cost \cite{mekhzoum}. Chen (2025) investigated supervised and semi-supervised point cloud learning for abnormality detection, showing improved robustness under limited annotations \cite{chen2025deep}. Yang and Gao (2025) presented PointCHD, a benchmark dataset for congenital heart disease classification and segmentation, highlighting the feasibility of point cloud learning in medical applications \cite{2025_Yang}.

However, these approaches primarily focus on tasks such as anatomical reconstruction, registration, and general abnormality detection rather than disease-specific classification in brain MRI. Moreover, they do not incorporate anatomically guided sampling or region-aware feature modelling, limiting their ability to capture neurodegenerative patterns relevant to Alzheimer’s disease.

In summary, existing medical point cloud approaches mainly focus on tasks such as segmentation, registration, and anatomical reconstruction, with limited application to brain MRI–based Alzheimer’s disease classification. These methods typically assume uniform importance across regions and lack disease-specific anatomical priors, while many also rely on computationally intensive architectures. In contrast, the proposed framework integrates anatomically guided sampling with region-aware feature learning to enable efficient and disease-specific Alzheimer’s disease classification.

\begin{figure*}[!t]
    \centering
    \includegraphics[width=\textwidth]{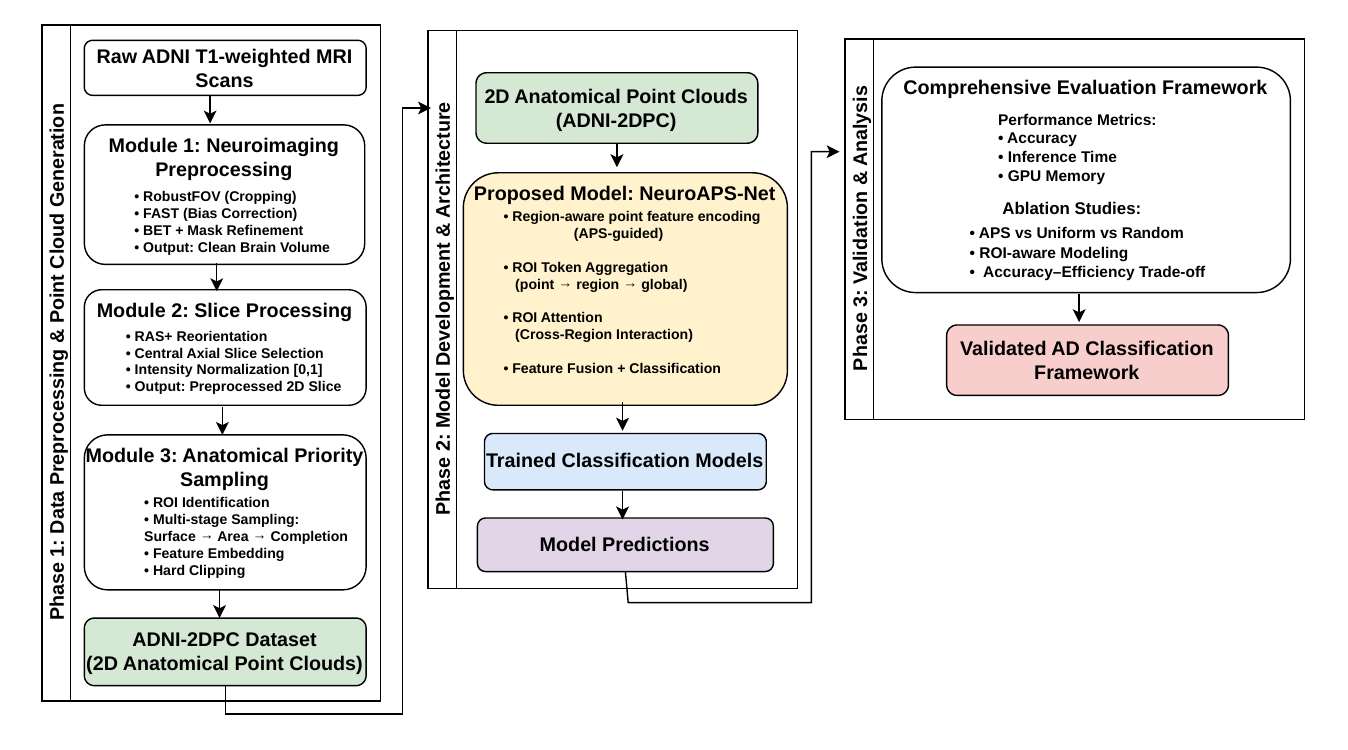}
    \caption{Overview of the proposed APS-guided 2D point cloud generation pipeline and NeuroAPS-Net architecture for Alzheimer’s disease classification.}
    \label{fig:neuroaps_net}
\end{figure*}

\section{Methodology}
\label{sec-method}
\subsection{Overall Framework}
This paper proposes a point cloud generation framework with a lightweight geometric deep learning architecture for Alzheimer’s disease classification, designed for deployment in resource-constrained clinical settings. The framework consists of three main stages: (1) neuroimaging preprocessing and anatomically informed point cloud generation, (2) the NeuroAPS-Net architecture for classification, and (3) training and inference for AD/CN prediction.

As shown in Figure~\ref{fig:neuroaps_net}, MRI scans are preprocessed, converted into anatomically informed 2D point clouds using APS, and classified by NeuroAPS-Net.

\subsection{Neuroimaging Preprocessing and Slice Generation}
Each T1-weighted MRI scan undergoes standard preprocessing to ensure consistency, including field-of-view cropping, bias-field correction using FAST, and brain extraction with mask refinement validated using FSL-BET, resulting in a clean brain-only volume. To reduce computational redundancy while preserving discriminative information, representative 2D axial slices are extracted to capture hippocampal, ventricular, and cortical morphology relevant to Alzheimer’s disease. All slices are reoriented to RAS+ and normalised to $[0, 1]$, forming the basis for anatomical sampling.

\subsection{Anatomical Priority Sampling for Point Cloud Generation}
\begin{figure}[t]
  \centering
  \includegraphics[width=1\columnwidth]{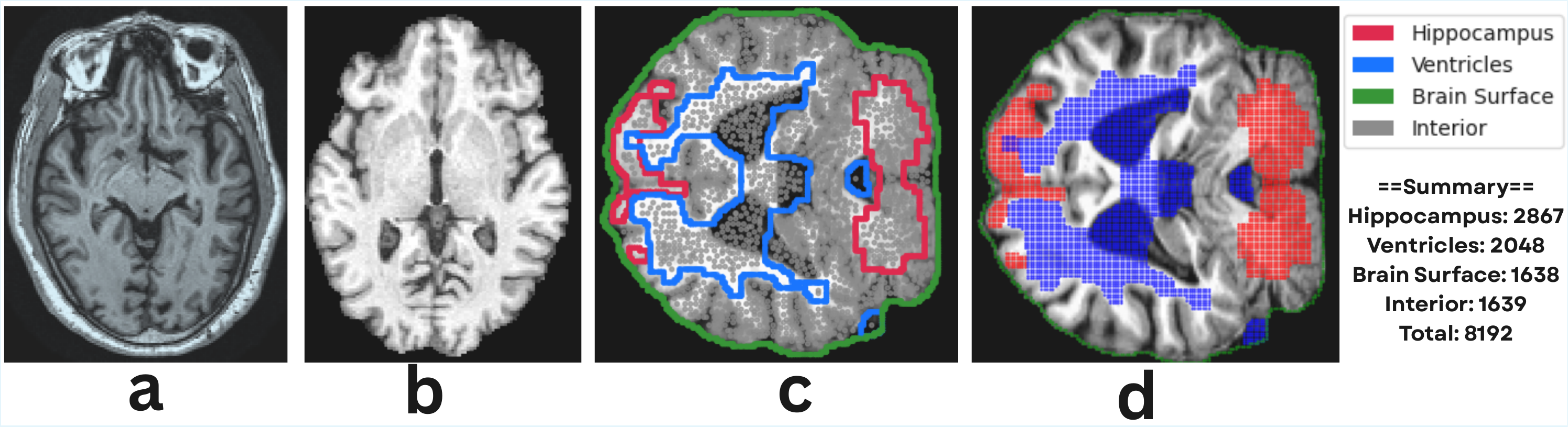}
  \caption{Anatomically informed point cloud generation pipeline using APS approach (a-d).
  (a) Raw ADNI T1-weighted MRI scan.
  (b) Preprocessed slice.
  (c) APS sampling of disease-relevant regions.
  (d) Region masks for fixed-size point generation.}
  \label{fig:aps_pipeline}
\end{figure}

Given a preprocessed 2D brain slice, APS is applied to generate a point cloud with emphasis on Alzheimer’s disease–relevant regions. Four anatomical categories are defined with fixed region-specific sampling ratios, as illustrated in Figure~\ref{fig:aps_pipeline}.

ROI identification is performed using anatomical masks or atlas-based segmentation. Unlike uniform sampling, APS allocates a predefined number of points to each region, ensuring representation of clinically significant structures such as the hippocampus. Sampling is conducted in multiple stages, including surface-aware sampling for cortical and ventricular boundaries, interior completion to preserve structural context, and hard clipping to remove points outside the brain mask.
Each sampled point is represented as, 
\begin{equation}
p_i = (x_i, y_i, I_i, r_i)
\end{equation}
where $(x_i, y_i)$ denotes spatial coordinates, $I_i$ is the normalised MRI intensity, and $r_i$ is the anatomical region label. The end result is a fixed-size, anatomically labelled 2D point cloud representing a single sample from the ADNI-2DPC dataset. The effectiveness of APS is evaluated through ablation studies in Section IV-D.

\subsection{NeuroAPS-Net Architecture}
\begin{figure*}[t]
    \centering
    \includegraphics[width=\textwidth]{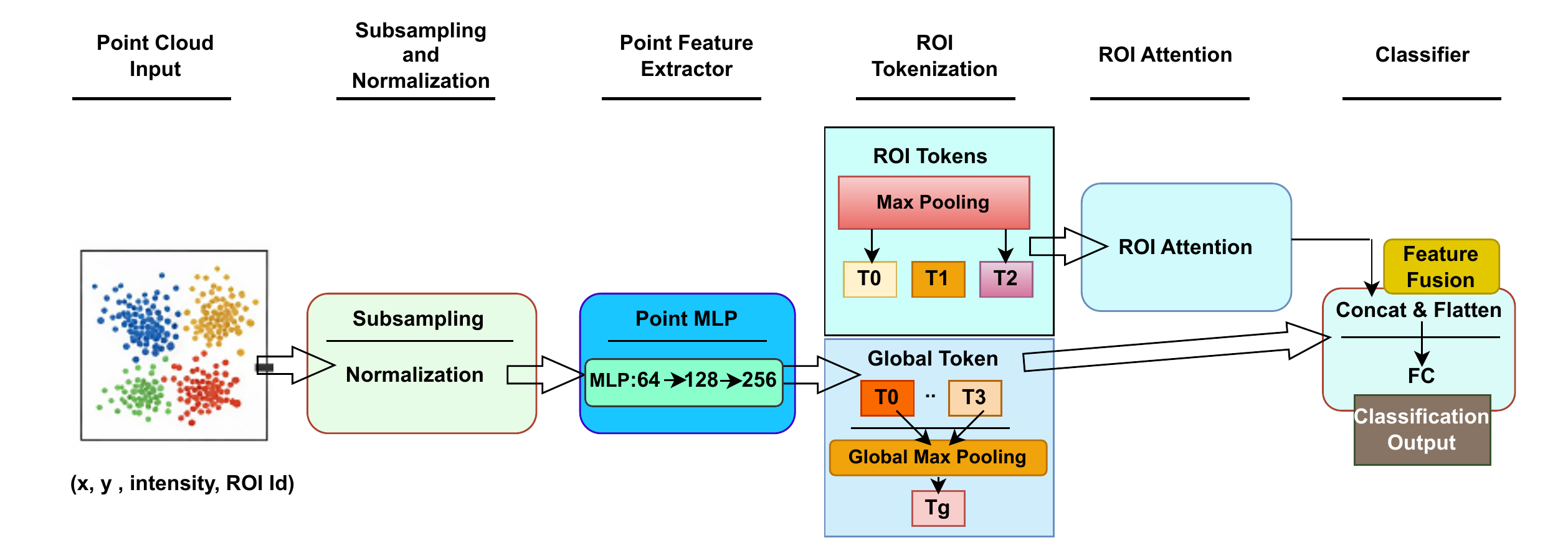}
    \caption{Overview of the proposed lightweight anatomically aware NeuroAPS-Net architecture}
    \label{fig:model_archi}
\end{figure*}

NeuroAPS-Net is a lightweight geometric deep learning architecture designed to exploit anatomically informed point cloud representations for deployment in resource-constrained clinical settings, as shown in Figure~\ref{fig:model_archi}.

\subsubsection{Point Feature Encoding}
Given an input point cloud $\{p_i\}_{i=1}^{N}$, a shared point-wise multilayer perceptron (MLP) is applied to extract point-level features:

\begin{equation}
f_i = \phi(p_i)
\end{equation}

where $\phi(\cdot)$ indicates an MLP with output dimensions $64 \rightarrow 128 \rightarrow 256$ and ReLU activations. This stage encodes spatial, intensity, and anatomical information into a unified feature representation.

\subsubsection{ROI Tokenisation via Region-wise Pooling}

To incorporate neuroanatomical knowledge, point features are grouped by ROI labels. For each region $r$, a token is obtained via max pooling:

\begin{equation}
T_r = \max_{i \,|\, r_i = r} f_i
\end{equation}

A global token $T_q$ is similarly derived across all points, producing a compact set of region-specific and global representations.

\subsubsection{ROI-Aware Feature Fusion}

ROI tokens are integrated with global context using a lightweight ROI attention mechanism. The aggregated regional representation is concatenated with the global token and passed to a compact fully connected classifier for prediction.

This design enables efficient interaction between regional and global features without relying on expensive graph construction or full self-attention. Unlike conventional point cloud models that treat all points uniformly, the proposed framework integrates neuroanatomical priors into both representation and learning, enabling region-aware feature modelling for Alzheimer’s disease. The effectiveness of this design is validated through ablation analysis in Section IV-D.

\subsection{Classification and Training Objective}

The combined feature representation is fed into a fully connected classification head to predict Alzheimer’s disease versus a cognitively normal status. The proposed network is trained end-to-end using the cross-entropy loss:

\begin{equation}
\mathcal{L} = - \sum_{c} y_c \log(\hat{y}_c)
\end{equation}

where $y_c$ and $\hat{y}_c$ represent the ground-truth and predicted class probabilities, respectively.

Due to its linear complexity with respect to the number of points and the absence of neighbourhood graph operations, NeuroAPS-Net achieves low inference latency and reduced GPU memory utilization, making it suitable for deployment in resource-constrained clinical settings.

\section{Experimental Results}
\label{ex-res}
\subsection{Experimental Setup, Dataset and Task Definition}
All models are implemented in PyTorch and evaluated on a system with an AMD Ryzen 1600 CPU, 16 GB RAM, and an NVIDIA RTX 3060 GPU (12 GB VRAM). This setup enables consistent benchmarking of inference latency and GPU memory usage. NeuroAPS-Net is trained end-to-end using the Adam optimizer with a fixed learning rate and cross-entropy loss. Lightweight data augmentation, including coordinate jittering and random point dropout, is applied to improve robustness.

All experiments use T1-weighted MRI data from the Alzheimer’s Disease Neuroimaging Initiative (ADNI), formulated as a binary classification task distinguishing Alzheimer’s disease (AD) from cognitively normal (CN) subjects. The dataset consists of 1000 subjects (500 AD, 500 CN) with a subject-level 80:20 train–test split, preserving class balance. The proposed preprocessing and APS pipeline converts each MRI scan into an anatomically informed 2D point cloud, forming the \textbf{ADNI-2DPC} dataset. Each sample consists of spatial coordinates, normalised intensity, and anatomical region labels (hippocampus, ventricles, brain surface, and interior). A fixed number of points is allocated per region via APS to ensure consistent anatomical coverage. ADNI-2DPC is designed to evaluate the efficient representation and learning of neuroanatomical MRI information using lightweight point cloud models, rather than to serve as a large-scale clinical benchmark.

\subsection{Baseline Models and Evaluation Metrics}

To provide a comprehensive comparison, NeuroAPS-Net is evaluated against widely used point cloud classification models, including PointNet, PointNet++, DGCNN, PointCNN, PointTransformer, CurveNet, SimpleView, and variants of PointCLIP. For a fair comparison, all baselines are trained and evaluated on the same input point clouds and data splits, with only minimal adaptation of the input layers to accommodate the proposed representation. Performance is evaluated using classification accuracy as the primary metric. In addition, inference latency (ms) and peak GPU memory consumption (MB) are reported to assess computational efficiency and deployment feasibility. While accuracy is the primary metric, additional metrics such as Precision, Recall, F1-score, and AUC-ROC are important for comprehensive evaluation in medical classification tasks and will be considered for future work.

\subsection{Experimental Results and Analysis}

\subsubsection{Overall Performance Comparison}
NeuroAPS-Net is evaluated against representative point cloud models using a fixed input size of 8192 points to ensure a consistent high-resolution comparison. Table~\ref{tab:overall_8192} reports classification accuracy, inference latency, and GPU memory consumption.

\begin{table}[t]
\centering
\caption{Overall performance comparison at 8192 points on ADNI-2DPC.}
\label{tab:overall_8192}
\begin{tabular}{c c c c}
\hline
\textbf{Model} & \textbf{Acc (\%)} & \textbf{Latency (ms)} & \textbf{GPU (MB)} \\
\hline
\hline
PointNet        & 83.33 & 2.17  & 543.0  \\
PointNet++      & 83.33 & 91.27 & 543.42 \\
DGCNN           & 81.82 & 135.72 & 6237.9 \\
PointCNN        & 80.30 & 119.56 & 800.74 \\
PointTransformer& 80.30 & 91.44 & 536.64 \\
PointCLIP       & 68.18 & 14.06 & 899.99 \\
PointCLIP V2    & 75.76 & 14.22 & 902.04 \\
CurveNet        & 74.24 & 3.74  & 6172.58 \\
SimpleView      & 74.24 & 1.54  & 1287.37 \\
\hline
\textbf{NeuroAPS-Net} & \textbf{84.85} & \textbf{1.48} & \textbf{234.6} \\
\hline
\end{tabular}
\end{table}

Under this setting, NeuroAPS-Net achieves the highest accuracy of 84.85\% while maintaining the lowest inference latency (1.48 ms) and GPU memory usage (234.6 MB). In contrast, graph-based models such as DGCNN incur high computational cost, requiring over 6 GB of GPU memory and more than 135 ms of latency despite lower accuracy. Transformer-based models and PointCNN similarly exhibit high latency and memory usage without performance gains. Projection-based methods such as SimpleView achieve competitive accuracy but require a large, fixed memory footprint, limiting scalability. CLIP-based models achieve low latency but consistently underperform in accuracy, suggesting a limited ability to capture subtle neuroanatomical variations in MRI-derived point clouds. These results demonstrate that \textbf{NeuroAPS-Net} achieves a strong balance between accuracy and computational efficiency, supporting its suitability for deployment in resource-constrained clinical settings.

\subsubsection{Effect of Point Density}

The sensitivity of NeuroAPS-Net to input resolution is evaluated at 2048, 4096, and 8192 points. Table~\ref{tab:point_density} summarizes classification accuracy, inference latency, and GPU memory usage across different models.

\begin{table}[t]
\centering
\caption{Effect of point density on classification accuracy, inference latency, and GPU memory usage.}
\label{tab:point_density}
\begin{tabular}{c c c c c}
\hline
\textbf{Model} & \textbf{Points} & \textbf{Acc (\%)} & \textbf{Latency (ms)} & \textbf{GPU (MB)} \\
\hline
\multirow{3}{*}{PointNet}
 & 2048 & 81.82 & 0.72 & 152.6 \\
 & 4096 & 83.33 & 1.28 & 282.7 \\
 & 8192 & 83.33 & 2.17 & 543.0 \\
\hline
\multirow{3}{*}{PointNet++}
 & 2048 & 81.81 & 87.69 & 161.54 \\
 & 4096 & 78.78 & 90.62 & 282.17 \\
 & 8192 & 83.33 & 91.27 & 543.42 \\
\hline
\multirow{3}{*}{DGCNN}
 & 2048 & 77.27 & 20.63 & 999.3 \\
 & 4096 & 77.27 & 51.72 & 1977.4 \\
 & 8192 & 81.82 & 135.72 & 6237.9 \\
\hline
\multirow{3}{*}{PointCNN}
 & 2048 & 71.21 & 93.12 & 288.33 \\
 & 4096 & 72.72 & 96.00 & 411.24 \\
 & 8192 & 80.30 & 119.56 & 800.74 \\
\hline
\multirow{3}{*}{PointTransformer}
 & 2048 & 66.67 & 91.03 & 504.77 \\
 & 4096 & 75.75 & 93.30 & 515.39 \\
 & 8192 & 80.30 & 91.44 & 536.64 \\
\hline
\multirow{3}{*}{PointCLIP}
 & 2048 & 63.63 & 13.91 & 899.99 \\
 & 4096 & 66.66 & 14.15 & 899.99 \\
 & 8192 & 68.18 & 14.06 & 899.99 \\
\hline
\multirow{3}{*}{PointCLIP V2}
 & 2048 & 65.15 & 14.17 & 902.04 \\
 & 4096 & 63.64 & 13.97 & 902.04 \\
 & 8192 & 75.76 & 14.22 & 902.04 \\
\hline
\multirow{3}{*}{CurveNet}
 & 2048 & 80.30 & 0.91 & 406.02 \\
 & 4096 & 72.72 & 1.28 & 1560.20 \\
 & 8192 & 74.24 & 3.74 & 6172.58 \\
\hline
\multirow{3}{*}{SimpleView}
 & 2048 & 77.27 & 1.54 & 1287.37 \\
 & 4096 & 84.84 & 1.75 & 1287.37 \\
 & 8192 & 74.24 & 1.54 & 1287.37 \\
\hline
\multirow{3}{*}{\textbf{NeuroAPS-Net}}
 & 2048 & 83.33 & 0.87 & 128.3 \\
 & 4096 & 83.33 & 1.21 & 163.7 \\
 & 8192 & \textbf{84.85} & \textbf{1.48} & \textbf{234.6} \\
\hline
\end{tabular}
\end{table}

As shown in Table~\ref{tab:point_density}, inference latency increases with point density across all models. For NeuroAPS-Net, latency scales approximately linearly, increasing from 0.87 ms (2048 points) to 1.48 ms (8192 points), reflecting its linear computational complexity and scalability. Even at the highest resolution, latency remains well within real-time clinical constraints. Again, increasing point density generally improves classification accuracy across models, indicating richer geometric and anatomical representation. However, performance gains vary, and some methods exhibit saturation or instability at higher resolutions. In contrast, NeuroAPS-Net shows consistent accuracy improvements as point counts increase, indicating effective utilization of additional points through anatomically guided sampling and region-aware feature learning.

These results highlight a clear trade-off between point density, classification performance, and computational cost. While higher densities improve accuracy, NeuroAPS-Net achieves competitive performance even at moderate resolutions with significantly lower latency and memory usage, supporting its suitability for efficient deployment in resource-constrained clinical settings.

\subsection{Ablation Analysis of Sampling and Anatomical Priors}

To quantitatively validate the effectiveness of APS and anatomical priors, we conduct an ablation study comparing different sampling strategies and architectural configurations. Specifically, five variants are evaluated: (1) APS-guided NeuroAPS-Net, (2) uniform sampling with ROI-aware modelling, (3) uniform sampling without ROI priors, (4) random sampling with ROI assignment, and (5) random sampling without ROI bias.

\begin{table}[t]
\centering
\caption{Ablation study on sampling strategy and anatomical priors at 8192 points on ADNI-2DPC.}
\label{tab:aps_ablation}
\footnotesize
\setlength{\tabcolsep}{3pt} 
\begin{tabular}{c c c c}
\hline
\textbf{Variant} & \textbf{Acc (\%)} & \textbf{Latency (ms)} & \textbf{GPU (MB)} \\
\hline
\hline
Uniform Sampling + ROI-aware        & 78.79 & 1.91 & 159.7 \\
Uniform Sampling (No ROI prior)     & 80.30 & \textbf{0.60} & \textbf{152.9} \\
Random Sampling + ROI-aware        & 81.82 & 1.37 & 159.7 \\
Random Sampling (No ROI prior)      & 81.82 & 0.61 & 153.6 \\
\hline
\textbf{NeuroAPS-Net (APS-guided)}  & \textbf{84.85} & 1.48 & 234.6 \\
\hline
\end{tabular}
\end{table}

Table~\ref{tab:aps_ablation} reports the results at 8192 points. The proposed APS-guided NeuroAPS-Net achieves the highest accuracy of \textbf{84.85\%}, outperforming all alternatives, with the strongest competing variant reaching 81.82\% ($\approx3\%$ improvement). This demonstrates that APS improves classification performance over uniform and random sampling by better utilizing disease-relevant anatomical information. Uniform sampling with ROI-aware modelling achieves only 78.79\% accuracy, indicating that region-aware aggregation alone is insufficient without anatomically guided sampling. Similarly, removing ROI priors degrades performance even with the same point cloud, highlighting the importance of region-aware feature encoding. The consistent improvement of ROI-aware variants over non-ROI configurations further confirms the importance of region-level feature aggregation. The results also reveal an accuracy–efficiency trade-off. Variants without anatomical priors exhibit lower latency and GPU usage but consistently underperform in accuracy. Although the APS-guided model is not the fastest, it maintains low inference latency (1.48 ms) comparable to lightweight baselines, indicating that the additional overhead of APS and ROI-aware modelling is minimal relative to the performance gain.

Overall, the ablation study confirms that both APS-based sampling and ROI-aware feature aggregation contribute significantly to performance, and their combination provides the best balance between accuracy and efficiency. These findings offer quantitative support for the proposed design and replace earlier qualitative interpretations of component contributions.

\subsection{Discussion: Accuracy–Efficiency–Point Density Trade-off}
The results support the primary motivation of this work: structural MRI can be effectively represented as anatomically informed 2D point clouds, enabling lightweight geometric deep learning as an efficient alternative to voxel-based 3D CNNs. While 3D CNN methods often achieve higher accuracy, they rely on computationally intensive volumetric processing. In contrast, the proposed framework prioritises a balance between accuracy and efficiency. It is not intended to replace 3D CNNs but to provide an efficient alternative that captures disease-relevant anatomical features at a lower computational cost. The use of 2D slices introduces a trade-off between efficiency and full 3D anatomical context. Although some volumetric information is lost, the selected slices retain key regions relevant to Alzheimer’s disease. Unlike standard 2D CNNs (e.g., ResNet, EfficientNet), which operate on dense grids without explicit anatomical priors, the proposed approach leverages anatomically guided sampling and region-aware learning to focus on clinically relevant regions. The results highlight a clear trade-off among accuracy, computational efficiency, and point density, further supported by the ablation study in section IV-D.

As point density increases, classification accuracy generally improves due to detailed geometric and anatomical information, but at the cost of higher latency and memory usage. Notably, NeuroAPS-Net achieves competitive accuracy even at moderate densities, indicating that APS effectively captures disease-relevant information without requiring dense sampling. By combining APS with region-aware modelling, NeuroAPS-Net maintains stable accuracy gains while scaling near-linearly in latency and memory consumption. This balance makes the framework suitable for deployment in resource-constrained clinical settings. The ablation results further confirm that performance improvements arise not only from point density but also from anatomically guided sampling and ROI-aware feature aggregation.

While experiments are conducted on the ADNI dataset, a standard benchmark in Alzheimer’s research, broader generalisation remains important. The proposed APS pipeline and NeuroAPS-Net are not dataset-specific and can be applied to other MRI cohorts with similar preprocessing, which will be explored in future work. Finally, results are reported from single-run experiments. Future work will include multi-run evaluation using statistical measures (e.g., mean and standard deviation) to improve robustness, as well as analysis of learned feature representations to improve interpretability.

\section{Conclusion}
This paper presents an anatomically informed framework for Alzheimer’s disease classification that represents structural MRI as compact 2D point clouds and evaluates them using a lightweight geometric deep learning architecture. The proposed Anatomical Priority Sampling emphasises disease-relevant regions, enabling efficient representation while preserving clinically meaningful neuroanatomical information. Building on this, NeuroAPS-Net integrates ROI-aware feature encoding to enable effective, and interpretable classification. Experimental results demonstrate competitive accuracy while significantly reducing inference latency and GPU memory usage compared to voxel-based CNNs and graph-based point cloud models. Sensitivity analysis reveals a clear accuracy–efficiency trade-off, where anatomically guided sampling maintains strong performance even at moderate point densities. Ablation results further validate the importance of APS and ROI-aware modelling in achieving optimal performance. Future work will focus on cross-dataset validation and multi-run statistical evaluation to improve generalization and robustness. Overall, the proposed framework provides a practical and deployable alternative for neuroimaging-based Alzheimer’s disease classification in resource-constrained clinical settings.

\bibliographystyle{IEEEtran}
\bibliography{references}

\end{document}